\ifx\XeTeXversion\undefined\pdfoutput=1\fi 
\documentclass[runningheads]{llncs}

\usepackage{graphicx}
\usepackage{float}

\usepackage{array}
\usepackage{comment}
\usepackage[export]{adjustbox}
\usepackage{amsmath}
\usepackage{amssymb}
\usepackage{bm}

\usepackage{booktabs}

\usepackage[hidelinks]{hyperref}
\usepackage{xurl}
\begin{document}
\title{Evaluating Deep Multivariate Imputation
Models on Wearable Device Data}

\author{
Skye Goodman\inst{1} \and
Roussel Desmond Nzoyem\inst{1,2} \and
Leandro Junges\inst{1,3}\orcidID{0000-0003-1536-3579} \and
Peter Kissack\inst{3} \and
Yasser Qureshi\inst{3} \and
Amberly Brigden\inst{1} \and
Jeff Clark\inst{1}\orcidID{0000-0003-0118-3999} \and
Nawid Keshtmand\inst{1}\orcidID{0009-0008-5552-1395}
}

\authorrunning{S. Goodman et al.}

\institute{School of Engineering Mathematics and Technology, University of Bristol, UK \and Centre for AI Fundamentals, University of Manchester, UK \and
Centre for Systems Modelling and Quantitative Biomedicine, University of Birmingham, UK\\
\email{nawid.keshtmand@bristol.ac.uk}}

\maketitle 
\begin{abstract}
Wearable device data enables continuous health monitoring, but suffers from structured missingness: features sharing a physical sensor drop out together. Deep imputation methods such as BRITS and SAITS have seen limited evaluation on multimodal physiological data under realistic missingness, and existing benchmarks use random-point holdout protocols that incorrectly assume missingness is independent across features and time.
Using data from a person with epilepsy recorded on a Garmin smartwatch, we develop an evaluation protocol that mines contiguous missing-run templates from training data, stratifies them by per-feature gap-length quantiles, and injects them as block masks with preserved co-missingness structure. A matched training protocol exposing models to the same missingness distribution reduces BRITS's severe-gap MAE by 43\%, demonstrating the potential benefit of the proposed evaluation and training protocol within this single-participant dataset. We further extend BRITS with time-of-day encoding and a circadian harmonic channel.
No single model dominates: linear interpolation is optimal for slow-moving features over short gaps; extended BRITS achieves lower MAE on dynamic cardiac features in moderate and severe gaps; and SAITS better preserves the ground-truth distribution by Jensen-Shannon distance despite higher MAE. 
Ultimately, model rankings strongly depend on evaluation designs. By exposing how traditional evaluation methods obscure true model capabilities, our transferable protocol establishes critical steps towards developing better imputation strategies for future multi-sensor wearable datasets.


\end{abstract}
 
\section{Introduction}
Wearable devices such as smartwatches have moved from consumer wellness into longitudinal clinical monitoring, with the global wearable medical device market projected to exceed USD 200 billion by 2035~\cite{kumar2024wearable}. A prominent application is epilepsy management, which affects over 50 million people worldwide~\cite{who2024epilepsy} and carries a sudden unexpected death risk of around 1 in 1,000 per year~\cite{wartmann2024sudep}. \emph{Seizure-risk forecasting}, which predicts an individual's likelihood of a future seizure, offers a potential way to alleviate the anxiety and loss of independence caused by their unpredictable nature~\cite{grzeskowiak2021forecasting}. Recent work has demonstrated the feasibility of training such models on non-invasive multimodal wearable data~\cite{stirling2021forecasting}, with performance comparable to intracranial-EEG benchmarks~\cite{nasseri2025forecasting}.

This progress, however, depends on a precondition the field has largely side\-step\-ped: \emph{missingness} in the training data, often due to sensor failures, poor skin contact, and connectivity dropouts \cite{lederer2023fitbit}. 
Time series \emph{imputation}, which confronts this challenge by replacing missing values with plausible surrogates, is a critical preprocessing step that directly shapes what downstream models can learn. Driven by the rise of \emph{deep learning}, this area has experienced a surge in research interest, enabling numerous healthcare successes \cite{le2024missing,kazijevs2023deep}. That said, key questions on how to realistically evaluate imputation strategies remain largely unresolved.

State-of-the-art (SOTA) deep imputation methods such as BRITS~\cite{Cao2018} and SAITS~\cite{Du2023} have been benchmarked predominantly under a \emph{random-point holdout} protocol, which uniformly removes a fixed proportion of observed values and implicitly assumes missingness is independent across features and time. On wearable data, this assumption fails as features sharing a physical sensor commonly drop out together for sustained intervals, and gap-length distributions vary by orders of magnitude across channels. Such methodological failures leave the community unable to gauge the true capabilities of these models, thus stifling the development of stronger imputation strategies to support patients. Details of various imputation approaches are discussed further in Appendix \ref{appendix: related_work}. Although recent diffusion-based methods have demonstrated strong performance \cite{FGTI}, they typically incur substantially higher computational cost. Accordingly, this work focuses on evaluating widely used recurrent (BRITS) and transformer-based (SAITS) architectures under a realistic wearable missingness protocol rather than providing an exhaustive comparison of all deep imputation methods.

Using physiological signals recorded from a Garmin smartwatch worn by a person with epilepsy \cite{thompson2026digital},
this paper makes three contributions to improve the training and evaluation of SOTA imputation methods for wearable devices:
\begin{enumerate}
\item A \textbf{wearable-realistic evaluation protocol} that mines contiguous missing-run templates from training data, stratifies them by per-feature gap-length quantiles into typical/moderate/severe buckets, and injects them as block masks with preserved co-missingness structure.
\item A \textbf{matched training schedule} exposing BRITS and SAITS to the same missingness distribution at training and evaluation, with a severity curriculum. This reduces BRITS's severe-bucket MAE by 43\% over its original training objective.
\item A \textbf{temporal context extension} for BRITS comprising sine--cosine time-of-day encodings and per-feature 24-hour circadian harmonic channels refitted weekly, reducing heart-rate MAE by 11\% and sleep MAE by 29\% under severe missingness.
\end{enumerate}
\section{Multi-Sensor Wearable Dataset Exploration}
\label{sec:data}
\subsubsection{Data collection.} Nine physiological time series variables (Table~\ref{tab:run_length_quantiles}) were collected and extracted from a Garmin smartwatch via the Labfront platform~\cite{labfront} for one recruited participant with epilepsy~\cite{thompson2026digital} (Ethics ref: 17455). Generalisability is discussed in Section~\ref{sec:limitations}. Raw Garmin CSV files are concatenated, sorted by Unix timestamp, and resampled onto a 60-second grid using the median within a $\pm$30s bucket (maximum for cumulative steps).
Features with native sampling intervals exceeding one minute (e.g.\ bodyBattery at 3 minutes) are forward-filled over their native interval to avoid false missingness classification. 

\subsubsection{Descriptive analysis.} A preliminary analysis revealed several key missingness properties underlying the multi-sensor data.

\begin{figure}[t]
    \centering
    \includegraphics[width=0.65\textwidth]{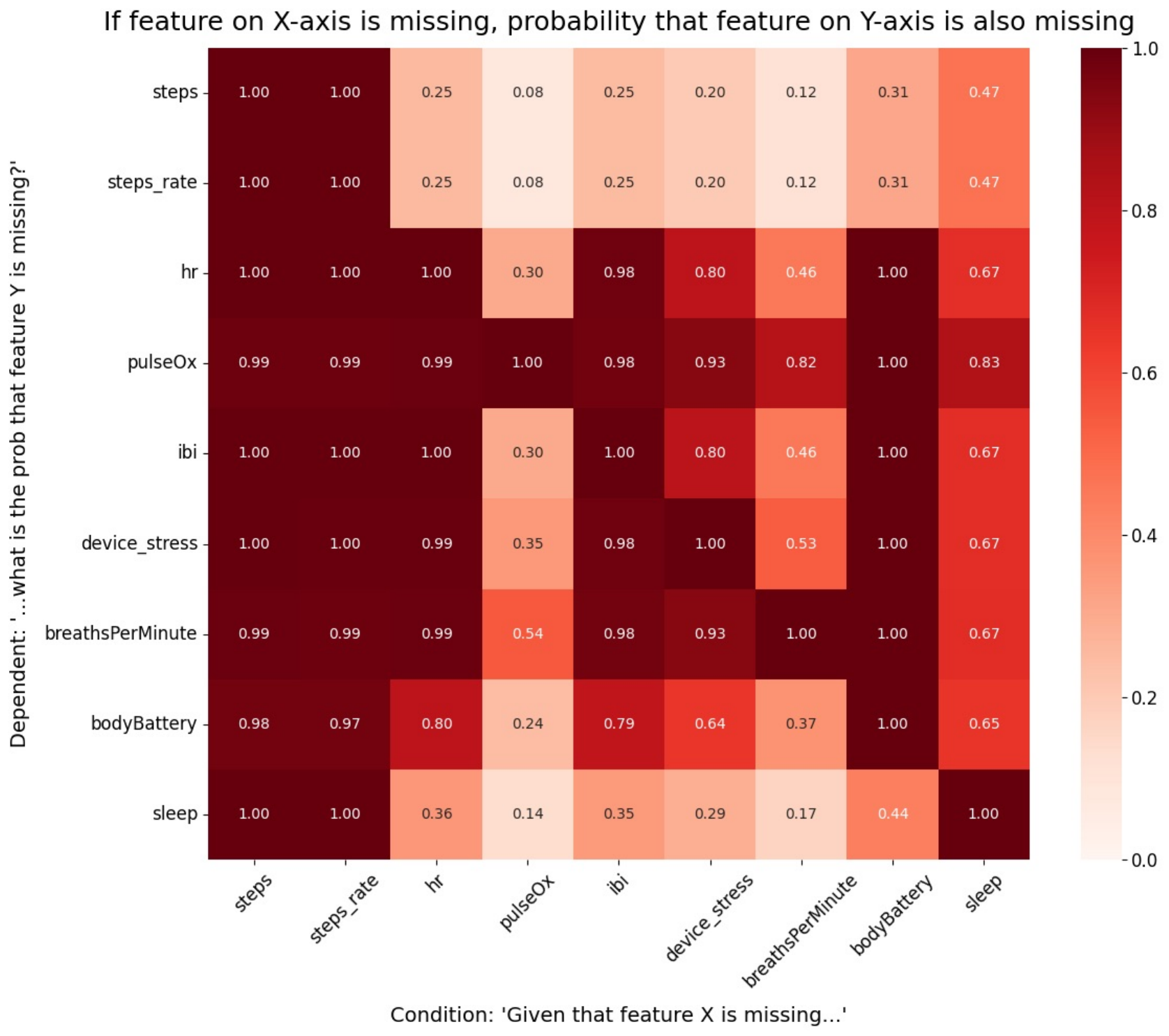}
    \caption{Conditional missingness matrix for the wearable features, where each cell reports $P(Y\ \mathrm{missing}\mid X\ \mathrm{missing})$. The heatmap highlights sensor-coupled co-missingness, including close to global feature dropouts when the accelerometer channels are missing, and a strongly coupled PPG-derived cluster of variables.}
    \label{fig:conditional-missingness}
\end{figure}

\paragraph{Sensor-coupled co-missingness.} The nine features derive from two sensors: (i) wrist optical photoplethysmography (PPG)~\cite{garmin_heart_rate_monitoring}, used for hr, ibi, pulseOx, device\_stress, and breathsPerMinute (with bodyBattery and sleep as downstream derived variables), (ii)  accelerometer, used for steps and steps\_rate. Shared hardware creates strong co-missingness (Fig.~\ref{fig:conditional-missingness}): $P(\text{ibi missing} \mid \text{hr missing}) = 1.00$, $P(\text{device\_stress missing} \mid \text{hr missing}) \approx 0.99$, and when the accelerometer fails, near-total dropout occurs across all channels. Co-missingness is not perfectly uniform within the PPG cluster: pulseOx has stricter signal-quality requirements and fails independently more often ($P(\text{hr } \allowbreak \text{missing} \mid \text{pulseOx missing}) \approx 0.30$).

\begin{table}[t]
\centering
\caption{Contiguous missing-run length statistics per feature, computed on the training set. A ``missing run'' is defined as one or more consecutive missing timestamps for a given feature. Run length is measured as the number of consecutive missing timestamps in the run. \#Runs reports the total number of missing episodes in the training data. $P_{50}$, $P_{75}$ and $P_{90}$ denote the 50th, 75th and 90th percentiles of run length $L$, and Max denotes the longest observed run in the training set.}
\label{tab:run_length_quantiles}
\scriptsize
\setlength{\tabcolsep}{4pt}
\begin{tabular}{llrrrrr}
\toprule
\textbf{Feature} & \textbf{Description} & \textbf{\#Runs} & $\boldsymbol{P_{50}}$ & $\boldsymbol{P_{75}}$ & $\boldsymbol{P_{90}}$ & \textbf{Max} \\
\midrule
steps            & Cumulative daily step count        & 37    & 56.0 & 105.0  & 212.4 & 236  \\
steps\_rate      & Interval-level step rate           & 38    & 53.0 & 101.25 & 209.8 & 236  \\
bodyBattery      & Garmin Body Battery score (5--100) & 83    & 35.0 & 98.0   & 207.8 & 1484 \\
hr               & Heart rate (bpm)                   & 162   & 25.0 & 63.75  & 154.7 & 1492 \\
ibi              & Inter-beat interval (ms)           & 212   & 18.5 & 39.25  & 120.1 & 1491 \\
sleep            & Sleep stage (0--4)                 & 107   & 16.0 & 39.0   & 185.0 & 1333 \\
device\_stress   & Garmin stress score (0--100)       & 1171  & 2.0  & 5.0    & 18.0  & 1492 \\
breathsPerMinute & Respiratory rate (breaths/min)     & 5368  & 2.0  & 3.0    & 7.0   & 1496 \\
pulseOx          & Peripheral oxygen saturation (\%)  & 12928 & 1.0  & 1.0    & 3.0   & 4389 \\
\bottomrule
\end{tabular}
\end{table}

\paragraph{Gap-length heterogeneity.} Contiguous missing-run statistics vary substantially across features (Table~\ref{tab:run_length_quantiles}). pulseOx is dominated by single-timestep dropouts, whereas hr, ibi, sleep, and bodyBattery exhibit much longer typical and upper-tail outages. A single uniform masking regime cannot capture this heterogeneity, motivating per-feature severity bucketing. 

\paragraph{Cross-feature signal under complete cases.} A pairwise Spearman analysis on complete cases identifies three correlation clusters: a near-deterministic cardiac pair (hr/ibi, $\rho = -0.99$); a sympathetic-activation group linking hr, ibi, and device\_stress ($\rho \approx 0.94$); and an activity--recovery group in which steps correlates positively with cardiac load and negatively with bodyBattery ($\rho \approx -0.47$). breathsPerMinute is weakly correlated with all other features ($|\rho| \le 0.14$). A shallow MLP trained per-target to predict each feature from the other eight at the same timestep confirms these clusters: hr and ibi are recovered with normalised MAE of 0.05--0.07, while pulseOx (0.68) and breathsPerMinute (0.66) are not. Crucially, these analyses are conducted under complete-case conditions, which may overstate the effect during real dropouts: the features with strongest pairwise predictability (hr/ibi) also drop out together with near-certainty, so the most informative cross-modal signal is largely unavailable during realistic missingness. This paradox motivates the wearable-realistic protocol below.
 
\section{Methodology}
\label{sec:method}
 
\subsection{Notation}
A multivariate window is $\mathbf{X} \in \mathbb{R}^{T \times D}$ with binary observation mask $\mathbf{M} \in \{0,1\}^{T \times D}$, where $m_{t,d}=1$ if $x_{t,d}$ is observed. For BRITS we additionally compute per-feature time-since-last-observation gaps $\delta_{t,d}$ recursively: $\delta_{t,d}=0$ if $m_{t,d}=1$, and $\delta_{t,d}=\delta_{t-1,d}+1$ otherwise, with boundary $\delta_{0,d}=0$. We adopt the original BRITS~\cite{Cao2018} and SAITS~\cite{Du2023} architectures unchanged for the forward pass. We use window length $T=360$ minutes throughout.
 
\subsection{Wearable-Realistic Block Masking}
\label{sec:wearable_realistic_masking}
 
For each target feature $d$, contiguous missing runs are mined from the training set. Each run of length $L$ contributes a multivariate template $\mathbf{M}_{\mathrm{run}} \in \{0,1\}^{L \times D}$ recording exactly which other features were also missing during the run. Templates are grouped by per-feature gap-length quantiles into three severity buckets:
\begin{align}
\text{typical}:\ L \le P_{50}^{(d)}, \quad
\text{moderate}:\ P_{50}^{(d)} < L \le P_{75}^{(d)}, \quad
\text{severe}:\ L > P_{75}^{(d)}.
\end{align}
At evaluation, a template is sampled and injected into a test window such that (i) the target is masked across the full block, (ii) helper features follow the template's co-missingness pattern, and (iii) the injection point satisfies a minimum fraction of ground-truth coverage in the target and a minimum agreement between the template's helper-availability pattern and the window's pre-existing missingness, ensuring physically plausible placements. For features whose maximum runs exceed the window length (e.g.\ pulseOx with runs up to 4389 timesteps), severe-bucket templates are truncated; the target may be missing for the entire window, forcing the model to impute purely from cross-feature and auxiliary context. 
Evaluation was stratified by feature and severity bucket rather than sampled according to the buckets' natural frequencies. For each feature, Typical, Moderate, and Severe were evaluated separately across the same set of test windows, with at most one template injected per window for each feature--bucket pair. Thus, the buckets had equal numbers of candidate windows, although the numbers of successful placements and evaluated timesteps could differ because run lengths vary and some placements may fail.
Importantly, per-feature reporting is retained because wearable features differ substantially in dynamics, missingness frequency, and gap lengths. 

Although severe gaps correspond to only the upper quartile of missing episodes by definition, they account for the majority of naturally missing timesteps in the training dataset across all features (60.7--88.5\%; e.g. hr 78.8\%, ibi 83.0\%, sleep 85.0\%), reflecting the long-tailed nature of wearable sensor outages. Consequently, improvements under severe missingness have substantial practical relevance despite the smaller number of severe missing episodes.
\subsection{Matched Training Schedule}
\label{sec:matched}

In the original BRITS objective \cite{Cao2018}, the feature-regression module, history regression, and fusion gate are supervised \emph{only} at naturally observed entries. In our dataset, 76\% of observed entries occur at timesteps where all eight other features are also observed, and 97\% where at least six are observed: the model is trained almost exclusively under near-complete helper availability and never sees the structured co-missingness conditions it faces at evaluation. The fusion gate in particular learns to down-weight the feature-regression estimate $\hat{\mathbf{x}}^f_t$ when helpers are substituted by historical estimates, a trade-off that the original training protocol rarely demands. SAITS already incorporates an MIT loss during training, but using random-point holdout that suffers the same train-to-test mismatch.

\begin{figure}[!h]
\centering
\includegraphics[width=0.8\textwidth]{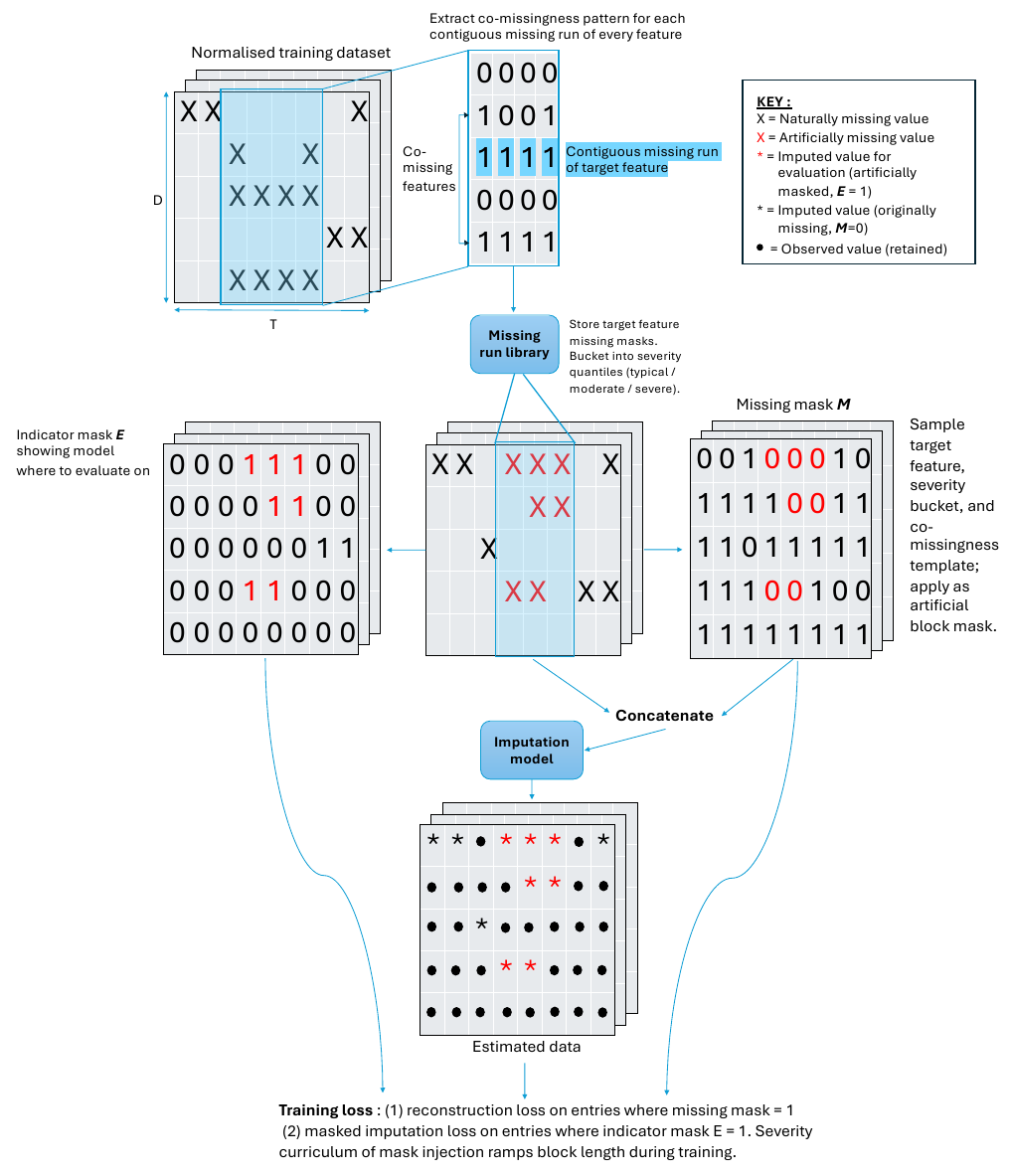}
\caption{Wearable-realistic training protocol used for BRITS and SAITS. Contiguous missing-run templates are extracted from training data and stored in a severity-bucketed library. During training, a template is sampled and injected into each window. Loss is computed on both naturally observed entries (reconstruction) and artificially held-out entries (masked imputation).}
\label{fig:training-schedule}
\end{figure}

We apply the same block masks described in Section~\ref{sec:wearable_realistic_masking} during training, producing an indicator mask $\mathbf{E} \in \{0,1\}^{T \times D}$ over artificially held-out positions (Fig.~\ref{fig:training-schedule}). The masked-reconstruction loss is:
\begin{equation}
\mathcal{L}_{\mathrm{mask}}
= \frac{\left\lVert \mathbf{E}\odot\bigl(\hat{\mathbf{X}}^{\mathrm{imp}}-\mathbf{X}\bigr)\right\rVert_1}
{\left\lVert \mathbf{E}\right\rVert_1 + \epsilon},
\label{eq:mask_loss}
\end{equation}
where $\hat{\mathbf{X}}^{\mathrm{imp}}$ is the model's imputation and $\epsilon = 10^{-8}$. For BRITS, masked reconstruction was not part of the original objective; we add it to the baseline loss as
\begin{equation}
\mathcal{L}_{\mathrm{total}}
= \mathcal{L}^{\mathrm{fwd}}_{\mathrm{imp}} + \mathcal{L}^{\mathrm{bwd}}_{\mathrm{imp}}
+ \lambda_{\mathrm{cons}}\,\mathrm{MAE}\!\left(\hat{\mathbf{X}}^{\mathrm{imp,fwd}},\hat{\mathbf{X}}^{\mathrm{imp,bwd}}\right)
+ \lambda(e)\,s\,\mathcal{L}_{\mathrm{mask}},
\label{eq:total_loss}
\end{equation}
where the first three terms are the baseline BRITS objective from~\cite{Cao2018} ($\lambda_{\mathrm{cons}}=10^{-1}$). $\lambda(e)$ is ramped from $\approx 0$ to $0.5$ over the early epochs so that masked reconstruction is introduced gradually. A per-batch scale factor $s = \mathrm{clip}(\mathcal{L}_{\mathrm{BRITS}}/\allowbreak(\mathcal{L}_{\mathrm{mask}}+\epsilon),\allowbreak\, 1,\allowbreak\, 2000)$ keeps the masked term on a comparable magnitude to the baseline objective and is detached from the computation graph. For SAITS, we retain the joint ORT$+$MIT objective of~\cite{Du2023} but substitute the random-point MIT mask with our block masks:

\begin{equation}
\mathcal{L}_{\mathrm{SAITS}}
=
\underbrace{
\tfrac{1}{3}\!\sum_{k=1}^{3}\!
\ell_{\mathrm{MAE}}(\tilde{\mathbf{X}}_k,\mathbf{X},\mathbf{M})
}_{\mathcal{L}_{\mathrm{ORT}}}
+
\underbrace{
\lambda\,\ell_{\mathrm{MAE}}(\hat{\mathbf{X}},\mathbf{X},\mathbf{E})
}_{\mathcal{L}_{\mathrm{MIT}}},
\end{equation}

with
$\ell_{\mathrm{MAE}}(\mathbf{Y},\mathbf{X},\mathbf{M}^{*})
=
\|\mathbf{M}^{*}\odot(\mathbf{Y}-\mathbf{X})\|_1
/
(\|\mathbf{M}^{*}\|_1+\epsilon)$,
where $\tilde{\mathbf{X}}_{1,2,3}$ are the intermediate SAITS reconstructions, $\hat{\mathbf{X}}$ is the final imputed output, and $\lambda=1$.

\paragraph{Severity curriculum.} Early epochs sample only typical templates; intermediate epochs add moderate; by two-thirds of training all three severities are sampled, with weights $(0.25, 0.30, 0.45)$ for typical/moderate/severe respectively. Both curriculum schedule and final weights were chosen on the validation split.

\subsection{Time-of-Day and Circadian Auxiliary Channels}
\label{sec:circadian}
 
Under severe co-missingness, helper availability collapses sharply for most target features, e.g.\ hr drops from 3.8 helpers (features that remain present when the target feature drops out) observed on average in the typical bucket to 2.1 in severe; steps and steps\_rate drop to near zero. To supply information that does not depend on sensor state, we append two channel types to the model input. First, a sine--cosine time-of-day encoding from the minute-of-day $n_t \in \{0,\dots,1439\}$,
\begin{equation}
\mathrm{tod}_{\sin}(t) = \sin\!\left(2\pi n_t/1440\right), \qquad \mathrm{tod}_{\cos}(t) = \cos\!\left(2\pi n_t/1440\right),
\end{equation}
which preserves the circular topology of clock time. Second, per-feature 24-hour harmonic channels fitted weekly by ordinary least squares on \emph{post-masking} observed entries (to prevent leakage):
\begin{equation}
\hat{y}_t = a + b\sin\!\left(2\pi \tau_t/1440\right) + c\cos\!\left(2\pi \tau_t/1440\right),
\end{equation}
evaluated at every timestep within the week. Coefficients $(a,b,c)$ are refit each week so the prior tracks gradual behavioural drift. Per-feature inclusion is governed by a Lomb--Scargle periodogram~\cite{VanderPlas2018LombScargle} analysis on the training set. bodyBattery (peak normalised power 0.58), sleep (0.51), ibi (0.34), hr (0.30), device\_stress (0.21), and pulseOx (0.13) show clear 24-hour periodicity and receive harmonic channels. sleep is excluded due to its four-state discrete nature; steps is excluded because its cumulative-then-reset pattern is poorly modelled by a smooth sinusoid; steps\_rate and breathsPerMinute lack sufficient 24-hour power. The choice of a single first-order harmonic over Natarajan et al.'s two-harmonic fit~\cite{Natarajan2025CircadianHeartRateActivity} reflects their finding that the 24-hour harmonic alone accounts for $\sim$85\% of the variance.
All auxiliary channels are concatenated to the feature dimension ($D=9 \to 16$). Their observation masks are fixed to 1, their BRITS time-gap encodings to 0, and they are excluded from all reconstruction losses and evaluation metrics: they are context, not imputation targets.
 
\section{Experimental}
\label{appendix:baselines}
We compare against two univariate classical baselines: last observation carried forward (LOCF) and linear interpolation (LI). BRITS uses recurrent hidden size $H=128$. SAITS uses $N=2$ DMSA blocks, model dimension $d_{\mathrm{model}}=256$, feed-forward dimension $d_{\mathrm{ffn}}=128$, $h=4$ attention heads with $d_k=d_v=64$, and dropout 0.1 (SAITS-base configuration of~\cite{Du2023}). Both models use batch size 64, training stride 60, evaluation stride 360 (non-overlapping), Adam optimisation with learning rate $10^{-3}$, and 400 steps per epoch. Hyperparameters were selected on the validation split.
 
\paragraph{Metrics.} We report mean absolute error (MAE) on standardised values
and symmetric mean absolute percentage error (sMAPE) in physical units after inverse-standardising. sMAPE is bounded in $[0\%, 200\%]$ and becomes unstable for zero-inflated features such as steps\_rate. For distributional fidelity we report Jensen--Shannon distance (JSDist) following~\cite{boursalie2022evaluation}:
\begin{equation}
    \mathrm{JSDist}(p, q) = \sqrt{\frac{\mathrm{KL}(p \,\|\, r)}{2} + \frac{\mathrm{KL}(q \,\|\, r)}{2}}, \qquad r = (p+q)/2,
\end{equation}
computed on a 50-bin discrete approximation on a shared support, bounded in $[0,1]$ and well-defined under non-overlapping support. All results are averaged over 10 seeds sharing identical sampled templates across all models, enabling matched-pair comparison.

\section{Results}
\label{sec:results}

\subsection{Effect of the Temporal Context Extension}
\label{sec:results-ext}
 
Table~\ref{tab:extension_comparison} compares baseline BRITS and SAITS using the nine features against the extension variants with time-of-day and circadian channels appended. 
For BRITS, the extension produces consistent improvements on cardiac and stress features across all severities, with the largest gains in severe (hr $-11\%$, ibi $-13\%$, sleep $-29\%$ despite sleep not receiving its own circadian channel). For SAITS, gains are smaller and inconsistent, with average MAE essentially unchanged in severe missingness. We attribute the asymmetry to architectural difference: SAITS's self-attention operates globally over the 6-hour window and can infer a physiological baseline from observed timesteps elsewhere in the window, partly substituting for what the circadian channel supplies. BRITS computes estimates locally; its hidden state degrades across long gaps and its feature regression has no access to non-adjacent observations. The auxiliary channels are most useful when locally available signal is most depleted. Based on these results we adopt \textbf{BRITS-ext} (with extension) and \textbf{SAITS} (without extension) as the strongest variants of each model in the remainder of this paper. 
In this proof of concept study the test set was used to determine which BRITS/SAITS extension variants to evaluate, future work should consider doing so on the validation set instead.
 
\begin{table}[t]
\centering
\caption{Effect of the temporal context extension on BRITS and SAITS under wearable-realistic block masking for the core nine features, and extended with time-of-day and circadian auxiliary channels appended. Each cell reports test MAE (mean $\pm$ std) and sMAPE (\%) per target feature and gap severity bucket over 10 seeded evaluation trials sharing identical sampled mask templates. Bold entries indicate the better variant within each model and row.
}
\label{tab:extension_comparison}
\tiny
\setlength{\tabcolsep}{3pt}
\renewcommand{\arraystretch}{1.15}
\begin{tabular*}{\textwidth}{@{\extracolsep{\fill}}lcccc@{}}
\toprule
\textbf{Feature} & \textbf{BRITS} & \textbf{BRITS-ext} & \textbf{SAITS} & \textbf{SAITS-ext} \\
\midrule
\multicolumn{5}{l}{\textit{Typical ($L \leq P_{50}$)}} \\
hr               & 0.386$\pm$0.017 / 5.93\%          & \textbf{0.366}$\pm$0.016 / 5.64\% & \textbf{0.377}$\pm$0.015 / 5.84\% & 0.385$\pm$0.016 / 5.92\% \\
ibi              & 0.280$\pm$0.017 / 5.12\%          & \textbf{0.276}$\pm$0.014 / 5.06\% & \textbf{0.296}$\pm$0.019 / 5.39\% & 0.298$\pm$0.016 / 5.44\% \\
device\_stress   & 0.268$\pm$0.019 / 17.47\%         & \textbf{0.254}$\pm$0.017 / 17.77\%& 0.309$\pm$0.022 / 20.01\%         & \textbf{0.292}$\pm$0.018 / 19.23\% \\
pulseOx          & \textbf{0.163}$\pm$0.021 / 0.45\% & 0.173$\pm$0.020 / 0.47\%          & 0.247$\pm$0.023 / 0.68\%          & \textbf{0.231}$\pm$0.028 / 0.63\% \\
steps            & \textbf{0.085}$\pm$0.008 / 69.7\% & 0.106$\pm$0.011 / 68.6\%          & \textbf{0.068}$\pm$0.006 / 68.9\% & 0.095$\pm$0.009 / 71.3\% \\
steps\_rate      & 0.171$\pm$0.034 / 67.2\%          & 0.171$\pm$0.034 / 67.2\%          & \textbf{0.170}$\pm$0.034 / 67.2\% & 0.171$\pm$0.034 / 67.2\% \\
bodyBattery      & \textbf{0.081}$\pm$0.006 / 5.71\% & 0.095$\pm$0.007 / 7.01\%          & \textbf{0.061}$\pm$0.003 / 3.32\% & 0.077$\pm$0.004 / 4.82\% \\
breathsPerMinute & \textbf{0.297}$\pm$0.019 / 8.26\% & 0.301$\pm$0.019 / 8.31\%          & \textbf{0.337}$\pm$0.021 / 9.34\% & 0.350$\pm$0.020 / 9.67\% \\
sleep            & \textbf{0.059}$\pm$0.011 / 8.41\% & 0.068$\pm$0.017 / 8.36\%          & \textbf{0.054}$\pm$0.014 / 6.96\% & 0.064$\pm$0.020 / 8.38\% \\
\textbf{Average} & \textbf{0.199} / 20.92\%          & 0.201 / 20.94\%                   & \textbf{0.213} / 20.84\%          & 0.218 / 21.41\% \\
\midrule
\multicolumn{5}{l}{\textit{Moderate ($P_{50} < L \leq P_{75}$)}} \\
hr               & 0.415$\pm$0.009 / 6.50\%          & \textbf{0.378}$\pm$0.008 / 5.92\% & 0.383$\pm$0.012 / 6.05\%          & \textbf{0.382}$\pm$0.010 / 5.99\% \\
ibi              & 0.349$\pm$0.012 / 6.27\%          & \textbf{0.319}$\pm$0.012 / 5.71\% & 0.330$\pm$0.011 / 5.91\%          & \textbf{0.329}$\pm$0.011 / 5.90\% \\
device\_stress   & 0.264$\pm$0.011 / 17.08\%         & \textbf{0.255}$\pm$0.010 / 17.64\%& 0.294$\pm$0.012 / 19.01\%         & \textbf{0.285}$\pm$0.014 / 18.98\% \\
pulseOx          & \textbf{0.163}$\pm$0.021 / 0.45\% & 0.173$\pm$0.020 / 0.47\%          & 0.247$\pm$0.023 / 0.68\%          & \textbf{0.231}$\pm$0.028 / 0.63\% \\
steps            & \textbf{0.100}$\pm$0.005 / 73.6\% & 0.121$\pm$0.004 / 72.9\%          & \textbf{0.085}$\pm$0.007 / 73.4\% & 0.114$\pm$0.005 / 73.0\% \\
steps\_rate      & 0.136$\pm$0.017 / 63.8\%          & 0.136$\pm$0.017 / 63.8\%          & 0.136$\pm$0.017 / 63.8\%          & 0.136$\pm$0.017 / 63.8\% \\
bodyBattery      & \textbf{0.106}$\pm$0.005 / 6.96\% & 0.120$\pm$0.007 / 8.36\%          & \textbf{0.081}$\pm$0.002 / 4.48\% & 0.086$\pm$0.003 / 5.33\% \\
breathsPerMinute & \textbf{0.317}$\pm$0.013 / 8.44\% & 0.322$\pm$0.012 / 8.55\%          & \textbf{0.327}$\pm$0.013 / 8.71\% & 0.349$\pm$0.018 / 9.32\% \\
sleep            & \textbf{0.134}$\pm$0.012 / 17.28\%& 0.139$\pm$0.016 / 16.58\%         & \textbf{0.103}$\pm$0.011 / 12.39\%& 0.112$\pm$0.014 / 12.78\% \\
\textbf{Average} & 0.220 / 22.27\%                   & \textbf{0.218} / 22.22\%          & \textbf{0.221} / 21.61\%          & 0.225 / 21.75\% \\
\midrule
\multicolumn{5}{l}{\textit{Severe ($L > P_{75}$)}} \\
hr               & 0.543$\pm$0.011 / 8.82\%          & \textbf{0.481}$\pm$0.009 / 7.76\% & 0.533$\pm$0.011 / 8.68\%          & \textbf{0.512}$\pm$0.011 / 8.34\% \\
ibi              & 0.489$\pm$0.015 / 8.43\%          & \textbf{0.426}$\pm$0.010 / 7.38\% & 0.480$\pm$0.013 / 8.25\%          & \textbf{0.452}$\pm$0.019 / 7.80\% \\
device\_stress   & 0.498$\pm$0.035 / 35.17\%         & \textbf{0.470}$\pm$0.032 / 33.87\%& \textbf{0.487}$\pm$0.027 / 35.02\%& 0.488$\pm$0.032 / 34.81\% \\
pulseOx          & 0.552$\pm$0.079 / 1.50\%          & \textbf{0.503}$\pm$0.067 / 1.37\% & \textbf{0.494}$\pm$0.053 / 1.35\% & 0.499$\pm$0.049 / 1.36\% \\
steps            & \textbf{0.137}$\pm$0.005 / 84.8\% & 0.141$\pm$0.003 / 82.8\%          & \textbf{0.091}$\pm$0.005 / 78.8\% & 0.131$\pm$0.005 / 80.1\% \\
steps\_rate      & 0.121$\pm$0.006 / 62.7\%          & 0.121$\pm$0.006 / 62.7\%          & 0.121$\pm$0.006 / 62.7\%          & 0.121$\pm$0.006 / 62.7\% \\
bodyBattery      & 0.336$\pm$0.021 / 16.03\%         & 0.336$\pm$0.015 / 16.84\%         & \textbf{0.354}$\pm$0.024 / 16.20\%& 0.362$\pm$0.027 / 16.78\% \\
breathsPerMinute & 0.435$\pm$0.021 / 11.16\%         & \textbf{0.418}$\pm$0.024 / 10.73\%& 0.469$\pm$0.032 / 12.12\%         & \textbf{0.434}$\pm$0.030 / 11.16\% \\
sleep            & 0.403$\pm$0.026 / 32.68\%         & \textbf{0.287}$\pm$0.015 / 27.02\%& \textbf{0.395}$\pm$0.024 / 31.34\%& 0.410$\pm$0.025 / 31.68\% \\
\textbf{Average} & 0.391 / 29.03\%                   & \textbf{0.354} / 27.83\%          & 0.380 / 28.28\%                   & \textbf{0.379} / 28.30\% \\
\bottomrule
\end{tabular*}
\end{table}

\subsection{Full Quantitative Comparative Study}
 
\begin{table}[!htb]
\centering
\caption{Comparison of BRITS-ext and SAITS against univariate classical baselines (LOCF - last observation carried forward, LI - linear interpolation) under wearable-realistic block masking. Each cell reports MAE (mean $\pm$ std) / sMAPE (\%). Bold indicates the lowest MAE in each row.}
\label{tab:baseline_comparison}
\tiny
\setlength{\tabcolsep}{3pt}
\renewcommand{\arraystretch}{1.15}
\begin{tabular*}{\textwidth}{@{\extracolsep{\fill}}lcccc@{}}
\toprule
\textbf{Feature} & \textbf{LOCF} & \textbf{LI} & \textbf{BRITS-ext} & \textbf{SAITS} \\
\midrule
\multicolumn{5}{l}{\textit{Typical ($L \leq P_{50}$)}} \\
hr               & 0.527$\pm$0.021 / 8.10\%          & 0.394$\pm$0.021 / 6.14\%          & \textbf{0.366}$\pm$0.016 / 5.64\% & 0.377$\pm$0.015 / 5.84\% \\
ibi              & 0.415$\pm$0.028 / 7.50\%          & 0.324$\pm$0.019 / 5.85\%          & \textbf{0.276}$\pm$0.014 / 5.06\% & 0.296$\pm$0.019 / 5.39\% \\
device\_stress   & 0.409$\pm$0.028 / 24.95\%         & 0.336$\pm$0.024 / 20.89\%         & \textbf{0.254}$\pm$0.017 / 17.77\%& 0.309$\pm$0.022 / 20.01\% \\
pulseOx          & 0.148$\pm$0.017 / 0.41\%          & \textbf{0.129}$\pm$0.015 / 0.35\% & 0.173$\pm$0.020 / 0.47\%          & 0.247$\pm$0.023 / 0.68\% \\
steps            & 0.042$\pm$0.012 / 15.56\%         & \textbf{0.021}$\pm$0.006 / 10.44\%& 0.106$\pm$0.011 / 68.6\%          & 0.068$\pm$0.006 / 68.9\% \\
steps\_rate      & 0.355$\pm$0.067 / 78.37\%         & 0.256$\pm$0.040 / 78.16\%         & 0.171$\pm$0.034 / 67.2\%          & \textbf{0.170}$\pm$0.034 / 67.2\% \\
bodyBattery      & 0.058$\pm$0.008 / 2.40\%          & \textbf{0.012}$\pm$0.001 / 0.51\% & 0.095$\pm$0.007 / 7.01\%          & 0.061$\pm$0.003 / 3.32\% \\
breathsPerMinute & 0.384$\pm$0.031 / 10.48\%         & \textbf{0.299}$\pm$0.021 / 8.32\% & 0.301$\pm$0.019 / 8.31\%          & 0.337$\pm$0.021 / 9.34\% \\
sleep            & 0.062$\pm$0.028 / 7.34\%          & \textbf{0.038}$\pm$0.015 / 4.69\% & 0.068$\pm$0.017 / 8.36\%          & 0.054$\pm$0.014 / 6.96\% \\
\textbf{Average} & 0.267 / 17.25\%                   & \textbf{0.201} / 15.04\%          & \textbf{0.201} / 20.94\%          & 0.213 / 20.84\% \\
\midrule
\multicolumn{5}{l}{\textit{Moderate ($P_{50} < L \leq P_{75}$)}} \\
hr               & 0.596$\pm$0.029 / 9.22\%          & 0.429$\pm$0.015 / 6.74\%          & \textbf{0.378}$\pm$0.008 / 5.92\% & 0.383$\pm$0.012 / 6.05\% \\
ibi              & 0.479$\pm$0.014 / 8.66\%          & 0.362$\pm$0.014 / 6.46\%          & \textbf{0.319}$\pm$0.012 / 5.71\% & 0.330$\pm$0.011 / 5.91\% \\
device\_stress   & 0.507$\pm$0.010 / 29.77\%         & 0.423$\pm$0.018 / 25.31\%         & \textbf{0.255}$\pm$0.010 / 17.64\%& 0.294$\pm$0.012 / 19.01\% \\
pulseOx          & 0.148$\pm$0.017 / 0.41\%          & \textbf{0.129}$\pm$0.015 / 0.35\% & 0.173$\pm$0.020 / 0.47\%          & 0.247$\pm$0.023 / 0.68\% \\
steps            & 0.079$\pm$0.008 / 25.98\%         & \textbf{0.043}$\pm$0.003 / 18.88\%& 0.121$\pm$0.004 / 72.9\%          & 0.085$\pm$0.007 / 73.4\% \\
steps\_rate      & 0.346$\pm$0.049 / 77.00\%         & 0.264$\pm$0.029 / 77.98\%         & \textbf{0.136}$\pm$0.017 / 63.8\% & \textbf{0.136}$\pm$0.017 / 63.8\% \\
bodyBattery      & 0.153$\pm$0.006 / 6.41\%          & \textbf{0.033}$\pm$0.002 / 1.46\% & 0.120$\pm$0.007 / 8.36\%          & 0.081$\pm$0.002 / 4.48\% \\
breathsPerMinute & 0.411$\pm$0.018 / 10.95\%         & \textbf{0.321}$\pm$0.015 / 8.59\% & 0.322$\pm$0.012 / 8.55\%          & 0.327$\pm$0.013 / 8.71\% \\
sleep            & 0.155$\pm$0.019 / 17.11\%         & \textbf{0.092}$\pm$0.012 / 10.95\%& 0.139$\pm$0.016 / 16.58\%         & 0.103$\pm$0.011 / 12.39\% \\
\textbf{Average} & 0.319 / 20.66\%                   & 0.233 / 17.43\%                   & \textbf{0.218} / 22.22\%          & 0.221 / 21.61\% \\
\midrule
\multicolumn{5}{l}{\textit{Severe ($L > P_{75}$)}} \\
hr               & 0.735$\pm$0.022 / 11.78\%         & 0.608$\pm$0.021 / 9.85\%          & \textbf{0.481}$\pm$0.009 / 7.76\% & 0.533$\pm$0.011 / 8.68\% \\
ibi              & 0.693$\pm$0.028 / 12.06\%         & 0.538$\pm$0.023 / 9.27\%          & \textbf{0.426}$\pm$0.010 / 7.38\% & 0.480$\pm$0.013 / 8.25\% \\
device\_stress   & 0.721$\pm$0.061 / 48.12\%         & 0.583$\pm$0.033 / 39.59\%         & \textbf{0.470}$\pm$0.032 / 33.87\%& 0.487$\pm$0.027 / 35.02\% \\
pulseOx          & 0.501$\pm$0.121 / 1.37\%          & \textbf{0.419}$\pm$0.099 / 1.14\% & 0.503$\pm$0.067 / 1.37\%          & 0.494$\pm$0.053 / 1.35\% \\
steps            & 0.180$\pm$0.010 / 52.00\%         & \textbf{0.078}$\pm$0.006 / 39.23\%& 0.141$\pm$0.003 / 82.8\%          & 0.091$\pm$0.005 / 78.8\% \\
steps\_rate      & 0.338$\pm$0.046 / 76.96\%         & 0.254$\pm$0.021 / 77.76\%         & \textbf{0.121}$\pm$0.006 / 62.7\% & \textbf{0.121}$\pm$0.006 / 62.7\% \\
bodyBattery      & 0.518$\pm$0.019 / 21.70\%         & 0.353$\pm$0.016 / 15.00\%         & \textbf{0.336}$\pm$0.015 / 16.84\%& 0.354$\pm$0.024 / 16.20\% \\
breathsPerMinute & 0.529$\pm$0.054 / 13.96\%         & 0.454$\pm$0.035 / 11.80\%         & \textbf{0.418}$\pm$0.024 / 10.73\%& 0.469$\pm$0.032 / 12.12\% \\
sleep            & 0.596$\pm$0.021 / 41.87\%         & 0.493$\pm$0.021 / 36.92\%         & \textbf{0.287}$\pm$0.015 / 27.02\%& 0.395$\pm$0.024 / 31.34\% \\
\textbf{Average} & 0.535 / 31.09\%                   & 0.420 / 26.73\%                   & \textbf{0.354} / 27.83\%          & 0.380 / 28.28\% \\
\bottomrule
\end{tabular*}
\end{table}

The performance gap between deep models and univariate baselines increases with gap severity (Table~\ref{tab:baseline_comparison}). LOCF is consistently the weakest approach, while LI remains competitive for slowly varying features and short gaps.

\paragraph{Cross-modal and dynamic features.} BRITS-ext performs best when strong cross-modal relationships or complex temporal dynamics can be exploited. The clearest example is device\_stress, which exhibits strong correlations with hr ($\rho=0.94$), ibi ($\rho=-0.94$), and sleep ($\rho=0.65$), while retaining high helper availability during missing periods (6.5 typical, 2.9 severe). In severe gaps, BRITS-ext reduces MAE from 0.583 (LI) to 0.470, a 19\% improvement. Its advantage also grows with severity for dynamic cardiac features: on hr, BRITS-ext improves over LI by 7\%, 12\%, and 21\% in the typical, moderate, and severe buckets, respectively.
Feature variability helps explain these results. The mean absolute first difference in standardised space is high for hr (0.243) and ibi (0.235), compared with just 0.004 for bodyBattery. Consequently, linear interpolation struggles on rapidly changing signals.



\paragraph{Where a classical baseline wins.} Linear interpolation (LI) outperforms BRITS-ext for bodyBattery and pulseOx across short and moderate gap lengths; BRITS-ext only wins during severe bodyBattery gaps. Because bodyBattery varies extremely slowly---its mean absolute first difference ($0.004$) is roughly 60 times smaller than the cardiac features---its short-to-moderate trajectory is nearly linear. Consequently, LI achieves near-perfect reconstruction (MAE $0.012$, sMAPE $0.51\%$) compared to BRITS-ext (MAE $0.095$, sMAPE $7.01\%$). At severe gaps, BRITS-ext usually outperforms LI however this is not true for the case of pulseOx which features a median first difference of zero, indicating prolonged stationarity, alongside weak cross-feature correlations ($|\rho| \le 0.26$). These factors allow LI to maintain superior performance even under severe gaps ($0.419$ vs. $0.503$). Together, these results suggest that feature-specific model selection would outperform a single imputation strategy applied uniformly.




\subsection{Effect of the Matched Training Protocol}

\begin{figure}[!htb]
\centering
\includegraphics[width=0.8\textwidth]{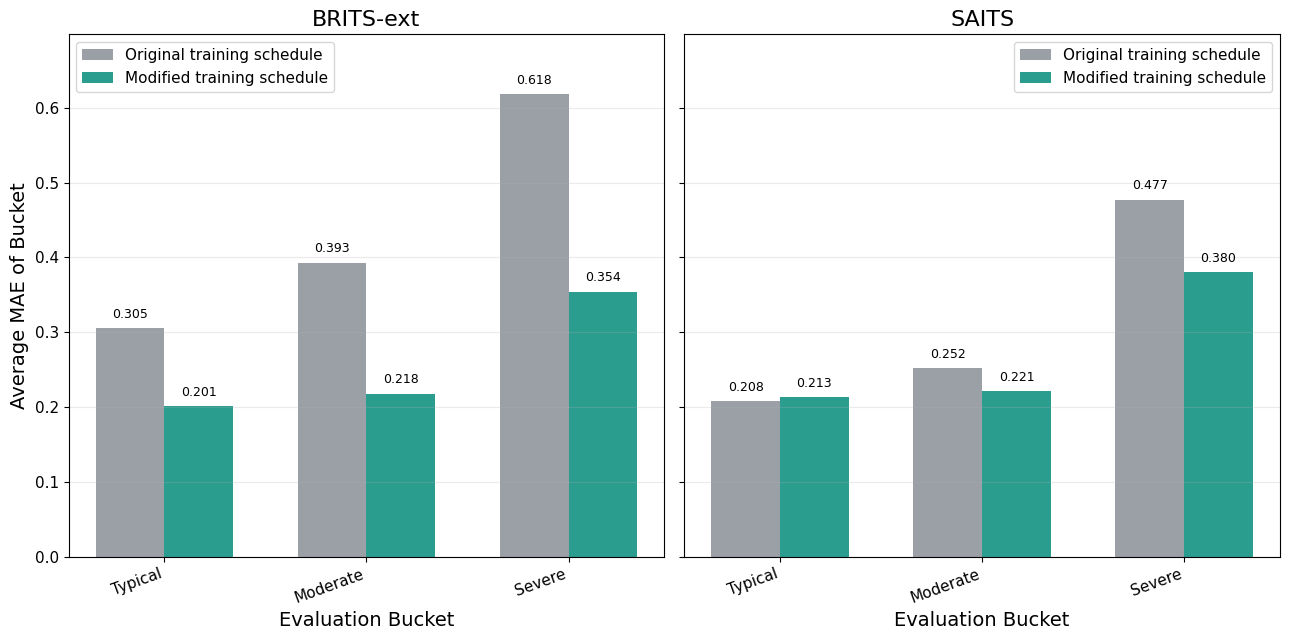}
\caption{Average bucket MAE for BRITS-ext and SAITS under the original training schedule (grey) versus the modified wearable-realistic block training schedule (green). Both variants are evaluated on the same masks, at the same positions, across the same 10 seeded trials. The only variable is the training objective. BRITS sees substantial gains across every bucket; SAITS gains more selectively, in the moderate and severe regimes.}
\label{fig:training_schedule_comparison}
\end{figure}

To isolate the contribution of the matched training schedule (Section~\ref{sec:matched}) from architectural and extension effects, we train each model under both its original schedule and ours, holding architecture, seeds, and evaluation masks fixed. For BRITS the original schedule is reconstruction on observed entries only; for SAITS the original schedule is the joint ORT
+ MIT objective with random-point MIT masking. Fig.~\ref{fig:training_schedule_comparison} reports the resulting average bucket MAE.

BRITS shows the largest effect of any result in this paper. Average bucket MAE falls by 34\% in typical (0.305 to 0.201), 45\% in moderate (0.393 to 0.218), and 43\% in severe (0.618 to 0.354). This is not a marginal gain: it is the difference between a model that fails under the dataset's actual missingness structure and one that performs the best on average across all models evaluated here. Under the original objective, BRITS's feature-regression and fusion-gate components are supervised almost entirely on near-complete helper availability, yet at evaluation they must operate under severe co-missingness. Training on the same block-masked distribution closes this train-to-test imbalance, with the largest gains concentrated in the severe bucket where the imbalance is greatest.

SAITS gains are smaller and more selective: essentially unchanged in typical (0.208 to 0.213), a 12\% improvement in moderate (0.252 to 0.221), and a 20\% improvement in severe (0.477 to 0.380). This is consistent with SAITS already incorporating an MIT loss under its original schedule; our protocol swaps one MIT mask distribution (random point) for another (wearable-realistic blocks). BRITS, by contrast, was seeing artificial masking during training for the first time.
A detailed feature-level analysis is provided in Appendix~\ref{sec:brits_ablation}.
\subsection{Distributional Fidelity}
While MAE captures pointwise accuracy, a model that consistently imputes near the feature mean can achieve low average error while failing to reproduce the distributional shape of the true signal~\cite{boursalie2022evaluation}. This matters for downstream seizure forecasting, where pre-seizure heart rate elevations manifest outside typical physiological ranges~\cite{bruno2018preictal}. Table~\ref{tab:jsdist} reports Jensen--Shannon distance (JSDist) between imputed and ground-truth distributions.
 
\begin{table}[!htb]
\centering
\caption{Jensen--Shannon distance (10-seed mean $\pm$ std) between imputed and ground-truth distributions, computed per feature and gap-severity bucket. Lower is better. Bold indicates the better-performing model for each feature and bucket.}
\label{tab:jsdist}
\tiny
\setlength{\tabcolsep}{3pt}
\renewcommand{\arraystretch}{1.15}
\begin{tabular*}{\textwidth}{@{\extracolsep{\fill}}lcccccc@{}}
\toprule
& \multicolumn{2}{c}{\textbf{Typical}} & \multicolumn{2}{c}{\textbf{Moderate}} & \multicolumn{2}{c}{\textbf{Severe}} \\
\cmidrule(lr){2-3}\cmidrule(lr){4-5}\cmidrule(lr){6-7}
\textbf{Feature} & \textbf{BRITS-ext} & \textbf{SAITS} & \textbf{BRITS-ext} & \textbf{SAITS} & \textbf{BRITS-ext} & \textbf{SAITS} \\
\midrule
hr               & 0.169$\pm$0.023          & \textbf{0.152}$\pm$0.015 & 0.163$\pm$0.009          & \textbf{0.142}$\pm$0.014 & 0.250$\pm$0.014          & \textbf{0.231}$\pm$0.008 \\
ibi              & \textbf{0.174}$\pm$0.016 & 0.178$\pm$0.016          & 0.157$\pm$0.013          & \textbf{0.143}$\pm$0.009 & 0.240$\pm$0.013          & \textbf{0.222}$\pm$0.016 \\
pulseOx          & \textbf{0.601}$\pm$0.058 & 0.689$\pm$0.031          & \textbf{0.601}$\pm$0.058 & 0.689$\pm$0.031          & \textbf{0.662}$\pm$0.033 & 0.693$\pm$0.039 \\
steps            & 0.244$\pm$0.025          & \textbf{0.219}$\pm$0.028 & 0.202$\pm$0.011          & \textbf{0.176}$\pm$0.010 & 0.192$\pm$0.009          & \textbf{0.148}$\pm$0.011 \\
steps\_rate      & 0.212$\pm$0.011          & \textbf{0.205}$\pm$0.011 & 0.181$\pm$0.007          & \textbf{0.181}$\pm$0.006 & \textbf{0.169}$\pm$0.003 & \textbf{0.169}$\pm$0.003 \\
device\_stress   & 0.249$\pm$0.024          & \textbf{0.231}$\pm$0.021 & 0.195$\pm$0.022          & \textbf{0.170}$\pm$0.012 & 0.212$\pm$0.017          & \textbf{0.196}$\pm$0.021 \\
bodyBattery      & 0.243$\pm$0.016          & \textbf{0.202}$\pm$0.021 & 0.198$\pm$0.015          & \textbf{0.148}$\pm$0.009 & \textbf{0.239}$\pm$0.010 & 0.253$\pm$0.018 \\
breathsPerMinute & \textbf{0.307}$\pm$0.022 & 0.329$\pm$0.022          & 0.329$\pm$0.020          & \textbf{0.316}$\pm$0.017 & 0.480$\pm$0.028          & \textbf{0.395}$\pm$0.035 \\
sleep            & \textbf{0.048}$\pm$0.017 & 0.054$\pm$0.021          & 0.114$\pm$0.019          & \textbf{0.046}$\pm$0.012 & 0.213$\pm$0.013          & \textbf{0.197}$\pm$0.016 \\
\textbf{Average} & \textbf{0.250}           & 0.251                    & 0.238                    & \textbf{0.223}           & 0.295                    & \textbf{0.278} \\
\bottomrule
\end{tabular*}
\end{table}
 
\begin{figure}[t]
\centering
\includegraphics[width=0.85\textwidth]{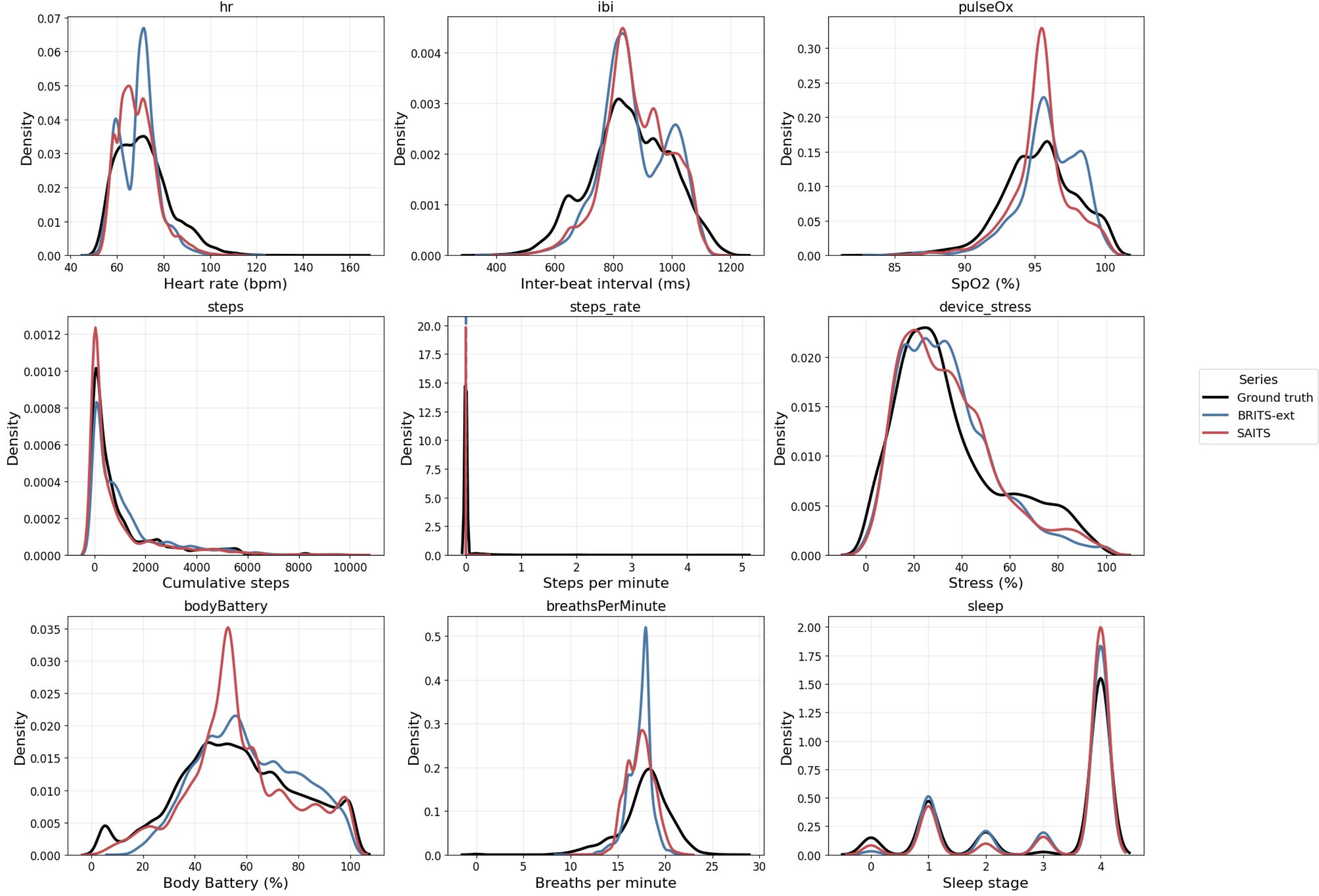}
\caption{Kernel density estimates of imputed values against ground-truth distributions at evaluation-mask positions, pooled across all three gap-severity buckets.}
\label{fig:kde_all_features}
\end{figure}

SAITS consistently outperforms BRITS-ext on JSDist for moderate and severe buckets, even on features where BRITS-ext shows higher pointwise MAE. This creates a tension with the MAE results (Table \ref{tab:baseline_comparison}): BRITS-ext produces more accurate predictions at individual timesteps, but SAITS better preserves the overall distributional shape. The kernel density estimates in Fig.~\ref{fig:kde_all_features} explain this. BRITS-ext over-sharpens distributional peaks, most visibly on hr (where its imputations exhibit a bimodal concentration not present in the ground truth) and on breathsPerMinute (where its central peak is exaggerated). Concentrating probability mass in narrow regions inflates JSDist even when those regions are near the correct values. Both models, however, share a clinically concerning failure: they suppress the upper tail of hr and both tails of ibi, precisely the regions where pre-seizure autonomic signal would manifest~\cite{bruno2018preictal}, and the regions downstream forecasting models would need an imputer to preserve. For sleep, both models reproduce the discrete-stage structure with sharp peaks, which is a result of the post-processing pipeline rounding outputs and clipping to $[0,4]$, not learned ordinal structure. Both over-weight the awake state (value 4), suggesting a default-to-most-common-class tendency during long gaps.
 JSDist and KDE are marginal distribution metrics; they complement rather than replace MAE, which remains the primary measure of pointwise fidelity.
 
\section{Limitations and Future work}
\label{sec:limitations}
This study has several limitations. First, KDE analysis showed that both deep models systematically under-represent the upper tail of hr and both tails of ibi. Because peri-ictal autonomic changes are expected to occur in these regions, low overall MAE may not translate into improved epilepsy forecasting performance. Second, the evaluation was conducted on data from a single participant. Missingness patterns and physiological baselines vary across individuals, limiting the extent to which the observed results can be generalised. Finally, end-to-end forecasting evaluation was beyond the scope of this work. Such an evaluation would require substantial computational resources and seizure labels with precise timing, whereas the dataset provides only a weekly self-reported seizure diary~\cite{brinkmann2021seizure}.

Several avenues for future work follow from these limitations. The feature-level results suggest that no single imputation model is optimal across all signals: linear interpolation performs best for slowly varying features and short gaps, whereas deep models consistently outperform baselines on more dynamic features. A hybrid framework that selects the imputation method on a per-feature basis using measurable characteristics such as signal variability and helper availability may therefore yield better overall performance. Future work should also investigate tail-aware training objectives, such as quantile losses or distribution-matching regularisers, to better preserve clinically relevant extremes. Finally, evaluating downstream forecasting performance and assessing cross-participant generalisation will be important for determining whether the observed imputation improvements translate into practical benefits.
\section{Conclusion}
 
We have shown that the random-point holdout protocol used to benchmark deep imputation models is misaligned with how wearable data actually fails. A wearable-realistic block-masking evaluation protocol, combined with a matched training schedule, changes the picture substantially: BRITS's severe-bucket MAE falls (improves) by 43\% when trained under the same missingness distribution it faces at evaluation. Under this evaluation no single model dominates. Linear interpolation remains best for slow-moving short-gap features; extended BRITS wins pointwise MAE on dynamic features; SAITS wins distributional fidelity. The practical implication is that no single imputation model should be deployed uniformly across a wearable feature set; the broader implication is that imputation model rankings depend strongly on how evaluation is designed. The protocol introduced here can be adapted to any multi-sensor wearable dataset for which contiguous-run statistics can be mined. To support reproducibility, the code used in this work is available at \url{https://github.com/skyegoodman/multivariate-wearable-imputation}.

\begin{credits}

\subsubsection{\ackname} RDN was supported by UKRI grant EP/S022937/1 (Interactive Artificial Intelligence), research startup funding from Professor Bahareh Tolooshams at the University of Alberta, and the UKRI Turing AI World-Leading Researcher Fellowship awarded to Professor Samuel Kaski. NK was supported by the EPSRC LEAP Digital Health Hub grant EP/X031349/1.

\subsubsection{\discintname}
The authors have no competing interests to declare that are
relevant to the content of this article.
\end{credits}

\bibliographystyle{unsrt}
\bibliography{references}
\newpage
\appendix

\section{Related Work}
\label{appendix: related_work}
\paragraph{Classical methods.} Mean/median imputation, LOCF, and linear/spline interpolation are efficient baselines with narrow applicability: linear interpolation suits slowly varying signals over short gaps but degrades with gap length or signal complexity. The Kalman filter~\cite{Welch1995} handles incomplete observations elegantly under linear-Gaussian dynamics, but these assumptions limit its use on nonlinear multimodal wearable signals. MICE~\cite{vanBuuren2011} and
k-nearest neighbours~\cite{Altman1992} were early attempts to exploit cross-feature structure.
\paragraph{Deep learning methods.} GRU-D~\cite{Che2018} introduced masking and time-decay inputs that reflect the intuition that older observations become less reliable, and that subsequent models build on. BRITS~\cite{Cao2018} extends this with bidirectional processing and a feature-wise regression module that exploits cross-feature correlations via a consistency loss between forward and backward passes. Diffusion-based approaches such as FGTI~\cite{FGTI} offer strong generative performance at substantial computational cost. SAITS~\cite{Du2023} uses two stacked diagonally-masked self-attention blocks with a learned weighted combination, trained jointly on observed reconstruction (ORT) and masked imputation (MIT) tasks. Foundation models such as GPT4TS~\cite{zhou2023fitsallpowergeneraltime}, Timer~\cite{liu2024timergenerativepretrainedtransformers}, and MOMENT~\cite{goswami2024momentfamilyopentimeseries} remain an active research direction, with open questions about transferability to physiological wearable signals~\cite{VAEs}.
\paragraph{The evaluation gap.} BRITS and SAITS are both evaluated under random-point holdout, which uniformly removes a fixed proportion of observed values. On ICU data this is defensible, as features arrive on heterogeneous schedules with largely feature-specific missingness. On wearables, most features share one of two physical sensors, making missingness highly sensor-coupled --- a structure that random-point holdout does not capture. Neither model has been evaluated on smartwatch data: BRITS trains solely on observed-value reconstruction, while SAITS introduces MIT with random-point masking. In Appendix B of~\cite{Du2023}, the authors note that applying MIT to BRITS yields mixed results across benchmarks, suggesting that training strategy interacts with both architecture and data characteristics --- but whether structured, realistic missingness patterns would alter these conclusions remains unexplored. The remainder of this paper addresses this gap.

\section{Training Protocol Ablation}
\label{sec:brits_ablation}

At feature level, cardiac features improve consistently for SAITS across all buckets (hr 0.693 to 0.533, ibi 0.638 to 0.480 in severe), and sleep roughly halves in moderate (0.245 to 0.103) and severe (0.639 to 0.395). Two features run against the trend: under the original schedule, SAITS achieves the lowest MAE of any deep model on pulseOx typical (0.156) and on device\_stress typical/moderate (0.217, 0.216). pulseOx's original-schedule severe MAE also sits slightly below the modified-schedule's (0.464 vs 0.494), and device\_stress inverts in severe (0.575 vs 0.487). pulseOx dropouts are mostly single-timestep events with near-full helper availability, so random-point masking already approximates them well; for pulseOx specifically there is little for the wearable-realistic protocol to fix. device\_stress points to a possible interaction between the severity curriculum's late-stage sampling weights and per-feature helper distributions. In the final phase of training, severity weights of 0.25/0.30/0.45 for typical/moderate/severe were chosen to concentrate signal on the hardest bucket; but severe masks also produce larger losses, and therefore larger gradient contributions, which late in training may pull the model's helper-weight calibration towards a regime where most helpers are absent. device\_stress has $\sim$6.5 helpers available in typical and moderate but $<$3 in severe, so a model calibrated for the severe regime may under-use helpers where they are plentiful. The original random-point schedule, by contrast, keeps helper availability close to the natural distribution throughout training, matching typical and moderate but leaving severe unsupervised; the two schedules therefore trade off performance across buckets. A natural extension would be to condition the model on a bucket indicator at training and test time so it can learn separate helper-use strategies for each regime rather than compromising across them.
\end{document}